\documentclass[manuscript,screen]{acmart}
\AtBeginDocument{%
  \providecommand\BibTeX{{%
    \normalfont B\kern-0.5em{\scshape i\kern-0.25em b}\kern-0.8em\TeX}}}

\setcopyright{acmcopyright}
\copyrightyear{2023}
\acmYear{2023}

\acmConference[SARs: TMI]{Conference on Human Factors in Computing Systems}{April 28, 2023}{Hamburg, Germany}
\acmBooktitle{1st Workshop on Socially Assistive Robots as Decision Makers: Transparency, Motivations, and Intentions at CHI 2023, April 23--28, 2023, Hamburg, DE} 
\acmPrice{0.00}
\acmISBN{978-1-4503-XXXX-X/18/06}

\begin{document}

\title{Communicating Complex Decisions in Robot-Assisted Therapy}


\author{Carl Bettosi}
\affiliation{%
  \institution{Edinburgh Centre for Robotics}
  \streetaddress{Heriot-Watt University, University of Edinburgh}
  \city{Edinburgh}
  \country{UK}}
\email{cb54@hw.ac.uk}

\author{Kefan Chen}
\affiliation{%
  \institution{Heriot-Watt University}
  \city{Edinburgh}
  \country{UK}}
\email{kc2039@hw.ac.uk}

\author{Ryan Shah}
\affiliation{%
  \institution{Heriot-Watt University}
  \city{Edinburgh}
  \country{UK}}
\email{r.shah@hw.ac.uk}

\author{Lynne Baillie}
\affiliation{%
 \institution{Heriot-Watt University}
 \city{Edinburgh}
 \country{UK}}
\email{l.baillie@hw.ac.uk}

%
%

\renewcommand{\shortauthors}{Trovato and Tobin, et al.}

\begin{abstract}
  Socially Assistive Robots (SARs) have shown promising potential in therapeutic scenarios as decision-making instructors or motivational companions. In human-human therapy, experts often communicate the thought process behind the decisions they make to promote transparency and build trust. As research aims to incorporate more complex decision-making models into these robots to drive better interaction, the ability for the SAR to explain its decisions becomes an increasing challenge. We present the latest examples of complex SAR decision-makers. We argue that, based on the importance of transparent communication in human-human therapy, SARs should incorporate such components into their design. To stimulate discussion around this topic, we present a set of design considerations for researchers.
\end{abstract}




\maketitle

\section{Introduction}


The recent advancements in assistive robotics aim to transform the way we deliver aspects of therapeutic care in areas such as physical and cognitive rehabilitation, behavioral therapy, and beyond. These technologies have been shown to provide promising results while having the potential to deliver significant social and economic benefits through improved access to care resources for communities~\cite{okamura2010medical} and decreased pressure on healthcare staff~\cite{qureshi2014impact}. In particular, the emerging field of Socially Assistive Robots (SARs) creates new social opportunities enabled by the robot's unique embodiment~\cite{kidd2004effect} allowing therapy experiences to be delivered closer to that of a human expert. Although the aim should not be to replace human therapists, there is now a wide body of research that explores how SARs can adopt instructor or companion-like roles in a supportive capacity to aid with therapy. Further investigation into this research has been encouraged by SARs' demonstrated ability to improve user engagement in therapeutic tasks when compared to the likes of non-embodied agents, such as social interactions facilitated through the use of a screen~\cite{lee2006physically,fasola2013socially, vasco2019train}.


In real-world human-to-human therapy sessions, experts take various decisions to maximise positive outcomes. In a physical rehabilitation scenario, for example, this may include adapting exercises to the patient's range of motion, employing different social behaviours such as encouraging or challenging utterances to boost engagement~\cite{ross2021observing}, or personalising to user traits to improve long-term relationships~\cite{tapus2008user}. Where human experts rely on trusted experience to do this, SARs must utilise complex decision-making processes. Early research in this area looked to use basic decision-making models such as rule-based systems~\cite{mataric2007socially, fasola2010robot}. However, the pursuit of highly personalised interaction combined with the advancements in sensory and computational technologies has increased the complexity of these decision-making processes.

This shift creates new challenges for SARs that look to communicate their decisions to users. In real-world therapy, it is natural for experts to justify decisions or answer patient concerns regarding the direction of therapy~\cite{wachtel1993therapeutic}. For complex decision-making processes such as machine learning models, developing explanations in high-stakes domains such as healthcare remains an open challenge~\cite{ghassemi2021false}. Additionally, failures or actions taken by SARs that lead to undesirable outcomes have shown to have a detrimental effect on trust~\cite{langer2019trust}. We stress that human therapists should always be on hand to intervene when patients are using assistive technology, however, such communication would be beneficial for healthcare staff also. In any case, transparent and understandable communication of decisions will be a vital component in future SAR-led therapy interventions.


Substantiated by a discussion of the current state-of-the-art, we discuss the impact of communication in complex SAR decision-making scenarios, focusing on therapeutic applications due to the sensitivity of autonomous decisions with regard to healthcare outcomes. Importantly, we discuss the communication of not the decision itself, but the process involved in calculating that decision. Moreover, we present key considerations for future SAR designers. Our aim is to raise awareness of this area and encourage the interaction design community to consider new and effective means of communicating complex decisions in sensitive and important applications such as therapy.

\section{Complex Decision-Making Processes in SARs}


The purpose of a SAR's decision-making component is to perceive input from the world and drive interactive behaviours. Therapy sessions can be considered as sequential decision-making processes, where past data can often influence a system's actions going forward~\cite{tsiakas2018task}. The mechanisms that drive these processes may use various techniques from basic rule-based approaches to complex machine-learning algorithms.

In ~\cite{gamborino2018interactive}, a SAR planning framework utilises Interactive Reinforcement Learning (IRL) to socially engage with and assist children in emotionally difficult situations. Experts interact with the children through the robot and, in turn, it learns to predict the next action of the expert based on the children's emotional response. ~\cite{clabaugh2019long} proposed a reinforcement learning-based SAR trained under a hierarchical human-robot learning framework to facilitate the social and educational development of children with autism spectrum disorder (ASD). The system allows the SAR to adapt and provide personalized feedback over long-term interaction with the user. The DREAM project proposes a supervised autonomous robotic system to improve robot-assisted therapy. Reinforcement learning and other supervised techniques (SPARC) allows the robot to learn from human demonstrations to improve future decision making, showing equivalent performance to traditional human-led therapies for children with ASD~\cite{cao2019robot}.



For cognitive therapy, ~\cite{hemminghaus2017towards} propose a reinforcement learning and multi-modal behaviour generation framework to assist users to complete games faster by learning how to guide their attention. Reinforcement learning is also used in ~\cite{tsiakas2018task} to provide feedback on engagement measured though physiological sensing, which results in more efficient real-time SAR personalisation for users engaging in games for cognitive improvement.


In the physical therapy/exercise domain, ~\cite{irfan2022personalised} presents a 2.5-year clinical study in which patients perform cardiac rehabilitation alongside a SAR companion which would provide motivational utterances based on task performance. Although aspects of interaction were based on some simple metrics like adherence and progress tracking, some decisions gathered and processed a complex array of sensor data in the environment. In an upper limb-focused scenario, ~\cite{pulido2019socially} presents a SAR which bases decisions on an automated planning mechanism, developed using Planning Domain Definition Language (PDDL). Firstly, therapists describe poses unique to each patient, a decision support system then builds a plan which the robot executes as sequential actions in the world and if actions fail, the plan regenerates. ~\cite{winkle2020situ} introduces an interactive machine learning approach in which expert therapists directly control actions on a SAR in response to patient actions in a real-world couch to 5K activity. State-action pairs are learned using an adapted K-means algorithm, which the SAR then uses to make autonomous decisions in future sessions. \cite{tapus2008user} investigates a reinforcement learning approach to optimise robot parameters to match introverted and extroverted personality types in a stroke rehabilitation scenario. Specifically, the agent manipulates proximity, speed and vocal content using task performance as a reward function.


Overall, there is a growing trend towards more sophisticated decision-making components in SAR-led therapy. Specifically, machine learning techniques such as reinforcement learning are frequently considered due to their suitability towards optimising future decisions based on long-term reward~\cite{akalin2021reinforcement}. In therapeutic scenarios which may last months to years, such approaches are highly relevant.

\section{Impact of transparent communication of decisions in SAR-Led Therapy}



When developing SARs, designers often look towards real-world interactions for insight~\cite{vsabanovic2014participatory}. In human-to-human therapy, as professionals become more regularly involved with patients, trust, assurance and feelings of security come as a result of better interactions and care involvement shaped to a patient's needs~\cite{kelly2015losing}. A key aspect of these interactions is an ability for experts to transparently communicate their real-time decisions in a reflective capacity~\cite{o2016influences}. Ultimately, a high level of transparency will not only improve interactions, but also
allow the therapist to learn from patient feedback. This knowledge to traverse complex decisions in
therapy settings comes with an evolution of competence along with years of training and experience provided by good feedback~\cite{jamarim2019nonverbal}.

Transparent communication on complex decision-making may play a pivotal role in building trust over long-term interactions, an area of growing importance in SAR-related research~\cite{langer2019trust}. Where human practitioners undergo extensive training and peer-review in strongly regulated environments, robots derive trust through the integrity of software that governs their complex decision making~\cite{shah2019privacy}. However, for non-system experts, such knowledge is not accessible. Therefore, transparent communication of decisions may help alleviate concerns around trust, whilst providing clearer insights into the robot's motivations and intentions.

In deployment, it is common that SARs experience failure. Such failures may result from technical errors due to sensing equipment for example, which can be notoriously difficult to recover from ~\cite{frennert2017case,cespedes2021socially}. Clear and transparent communication of robot failures can have a significant impact on user trust~\cite{nesset2021transparency}. Additionally, in scenarios where the robot has not failed but the interaction has arrived at an undesired state, for example, low engagement in therapy practice, communication of decisions may help the user to understand as to why the SAR took certain paths and, in turn, this transparency may improve the relationship.

Current research is interested in the adaption of SAR behaviour over multiple interactions to compliment the long-term nature of therapy. To achieve this, decision-making processes are utilising learning methods which require sufficient and accurate data that reflects the real-world state. Communication from SARs that can explain these processes could prompt feedback from the user on how well it is performing~\cite{schneider2021comparing}. This data could, in turn, be used to support the learning process~\cite{brys2015reinforcement}.

\section{Design Considerations for Effective Communication}

Given the potential impact of transparent communication in SAR-led therapy, we present a set of design considerations to enable future designers to begin thinking about this topic.

\textit{\textbf{What decisions will be communicated?}} Deciding upon the exact SAR decisions to be communicated will depend on the task at hand. For example, in cognitive rehabilitation, SARs have been seen to instruct users through memory games, adjusting levels of difficulty depending on task performance~\cite{tsiakas2018task}. In such a scenario, a justification of the change in difficulty may prove useful. Where users may question the decision-making process at any point during the interaction, designers should consider models which can be traced back to understand the sequence of activities that led to the decision. 

\textit{\textbf{Who will decisions be communicated to?}} To build effective communication of decisions, it is important to consider characteristics of the end-user such as: their acceptance and trust of robots, for example, cultural background~\cite{nomura2017cultural} and age~\cite{fridin2014acceptance} are seen to have an effect on acceptance and therefore may require different approaches in communication; level of interest in understanding decision-making; their ability to comprehend what is being explained. Healthcare staff who may often act as an interface between the technology and patient to support and motivate the SAR's use. For this reason, it will also be important to consider that the SAR may need to communicate decisions to this user base.

\textit{\textbf{When will decisions we communicated?}} Calculating the correct timing of decision communication may depend on a number of considerations such as: the significance of the decision or a change in the therapy direction (task difficulty increase/decrease or rest period); when feedback is provided by the user such as telling the SAR that this particular task is challenging or too easy; when questions arise from the patient or therapist and the SAR must provide rationale as to why decisions where made; when unexpected failure occurs and the SAR should find some reason as to why this happened; when unintended results are detected, such as low engagement from the user or poor therapeutic outcomes. Furthermore, designers should consider the delicate interplay between the communication of decisions and the communication of the explanation of these decisions, and striking a timely balance between both which does not overload the user with information.

\textit{\textbf{How will decisions be communicated?}} Expert therapist communication is actioned via verbal (e.g. speech or text) or non-verbal means (e.g. touch). Although touch can largely stimulate communication, assurance and security with patients~\cite{aredes2013comunicaccao,das1975,bjorbaekmo2016touch}, SARs are considered as hands-off assistive technologies which focus on the social interaction as opposed to the physical~\cite{mataric2016socially}. Verbal communications may provide a natural interaction closest to that of a real-world therapist, however, for SARs equipped with visual interfaces, the communication of complex decisions through visual means may prove effective~\cite{van2022correct}, which has been hinted to already in ~\cite{lee2020towards}. Overall, the consideration of how exactly decisions will be communication will largely depend on the complexity of the underlying process. Designers should carefully consider how to translate complex back-end processes to interpretable information that does not overwhelm the user.

Communication of SAR decisions is rarely considered in research yet is a key factor of real-world therapy. We argue that, as researchers strive for greater personalised and adaptive long-term solutions through the use of complex decision-making processes, they must consider how to effectively communicate these decisions. This transparency, we believe, will be essential to the trust and acceptance of these technologies. Towards this, we direct future designers to these design considerations. We focus on therapy due to the importance of transparent communication in human-human interaction, yet these design considerations will likely be relevant in a whole host of SAR applications.

\bibliographystyle{ACM-Reference-Format}
\bibliography{sample-base}

\appendix

\end{document}